\documentclass{article}

     \PassOptionsToPackage{numbers, compress}{natbib}

\usepackage[preprint]{neurips_2024}

\usepackage{placeins}
\usepackage[utf8]{inputenc} 
\usepackage[T1]{fontenc}    
\usepackage{url}            
\usepackage{booktabs}       
\usepackage{amsfonts}       
\usepackage{nicefrac}       
\usepackage{microtype}      
\usepackage{xcolor}         
\usepackage{amsmath}         
\usepackage{graphicx}
\usepackage{multirow}

\definecolor{thecolor}{RGB}{60, 117, 175}
\usepackage[citecolor=thecolor, linkcolor=thecolor,colorlinks=true, urlcolor=thecolor]{hyperref}       

\title{How Many Bytes Can You Take \\ Out Of Brain-To-Text Decoding?}

\author{%
  Richard J. Antonello \\
  Department of Computer Science\\
  The University of Texas at Austin\\
  \texttt{rjantonello@utexas.edu} \\
   \And
   Nihita Sarma \\
   Department of Computer Science \\
   The University of Texas at Austin \\\texttt{nihitasarma@utexas.edu} \\
   \AND
   Jerry Tang \\
   Department of Computer Science \\
   The University of Texas at Austin \\\texttt{jerrytang@utexas.edu} \\
   \And
   Jiaru Song \\
   Department of Computer Science \\
   The University of Texas at Austin \\\texttt{jiarus@cs.utexas.edu} \\
   \And
   Alexander G. Huth \\
   The University of Texas at Austin \\
   Departments of Computer Science and Neuroscience \\
   \texttt{huth@cs.utexas.edu} \\
}

\begin{document}

\maketitle

\begin{abstract}
Brain-computer interfaces have promising medical and scientific applications for aiding speech and studying the brain. In this work, we propose an information-based evaluation metric for brain-to-text decoders. Using this metric, we examine two methods to augment existing state-of-the-art continuous text decoders. We show that these methods, in concert, can improve brain decoding performance by upwards of 40\% when compared to a baseline model. We further examine the informatic properties of brain-to-text decoders and show empirically that they have Zipfian power law dynamics. Finally, we provide an estimate for the idealized performance of an fMRI-based text decoder. We compare this idealized model to our current model, and use our information-based metric to quantify the main sources of decoding error. We conclude that a practical brain-to-text decoder is likely possible given further algorithmic improvements.
\end{abstract}

\section{Introduction} 

Recent advancements in natural language processing research \cite{vaswani2017attention, radford2019language, zhang2022opt, touvron2023llama} have spurred developments in our ability to decode language from brain recordings \cite{sun2019towards, gauthier2019linking, sun2020brain2char, tang2023semantic, caucheteux2022deep, oota2023deep}. These \textit{semantic decoding models} are typically built either as direct decoders \cite{pereira2018toward, affolter2020brain2word} or as Bayesian decoders \cite{tang2023semantic, oota2023deep, zhang2024brain}. \textit{Direct decoders}, as their name implies, directly regress from brain recordings to a linguistic feature space \cite{affolter2020brain2word}. \textit{Bayesian decoders}, which we primarily use in this work, use Bayes' law to invert \textit{language encoding models}, which solve the problem of predicting neural recordings from a linguistic feature space such as the hidden states of a language model \cite{mitchell2008predicting, wehbe2014simultaneously, huth2016natural, naselaris2011encoding}. 

Bayesian semantic decoders for fMRI were introduced in \cite{tang2023semantic}, which established that it is possible to continuously predict the words that a person is hearing or thinking from fMRI recordings. This decoder is structured as a beam search that uses a language model to propose extensions to a beam of decoder predictions. The relative probabilities of observing the brain data given each extension are estimated using an encoding model and the most likely extensions are maintained on the beam. 

Brain decoding has been demonstrated for a variety of other domains including vision \cite{kay2008identifying, naselaris2009bayesian, nishimoto2011reconstructing, shen2019deep, takagi2023high} and motor volition \cite{willett2023high, lorach2023walking,anumanchipalli2019speech}. In language, brain decoding has been demonstrated using other recording modalities such as ECoG \cite{chen2024neural, chen2024subject}, EEG \cite{farwell1988talking, saeidi2021neural, defossez2023decoding, liu2024eeg2text, murad2024unveiling}, and MEG \cite{chan2011decoding, dash2019decoding}. Recently, brain decoders that use recordings from invasive techniques such as ECoG have become highly accurate, and have the potential to be used as speech neuroprostheses. However, these techniques require neurosurgical implantation of recording electrodes. In contrast, semantic decoders operate on signals that can be measured non-invasively using methods like functional MRI, which could eventually provide a cheaper and safer alternative.

Semantic decoding models have attained reasonable performance on continuous language, but in order to be useful as a neuroprosthetic device, the decoding accuracy must be excellent \cite{goering2021recommendations}. Yet these models are still very new, and there are likely significant algorithmic gains that can be made. Furthermore, evaluating these models remains challenging, as they tend to paraphrase text rather than producing word strings with exact matches. In this work, we establish a quantitative information-theory based metric for evaluating and comparing semantic decoding models across datasets and subjects. We identify two relatively simple methods for improving the performance of existing state-of-the-art decoding models\cite{tang2023semantic}. We show that these methods are capable of substantially improving semantic decoding models both qualitatively and using our new metric. We then use a noise ceiling analysis to explore the theoretical limits of semantic decoding models, providing an estimation for their potential performance and qualifying what is achievable at the spatiotemporal resolution provided by modern 3T fMRI. We conclude with a discussion of the practical consequences of an ideal or near-ideal decoder, as well as ethical considerations.

\section{Methods}

\subsection{Bayesian Decoding}

The goal of language decoding is to find the word sequence $S$ that maximizes the probability distribution $P(S|R)$ over word sequences given brain responses $R$. In Bayesian decoding, this can be done by using a voxel-wise encoding model (and multivariate normal noise model) to estimate $P(R|S)$ and a language model to estimate $P(S)$ \cite{tang2023semantic, naselaris2011encoding, nishimoto2011reconstructing}. 

A voxel-wise encoding model $\hat{R}$ is trained to predict brain responses $\hat{R}(S)$ from features of $S$ \cite{huth2016natural}. Assuming that BOLD signals are affected by Gaussian random noise, the probability distribution $P(R|S)$ can be modeled as a multivariate Gaussian distribution with mean $\mu = \hat{R}(S)$ and covariance $\Sigma = \langle (R - \hat{R}(S))^T(R - \hat{R}(S)) \rangle$ \cite{nishimoto2011reconstructing}. The covariance $\Sigma$ models the structure of correlated noise across voxels. In this study, stimulus features were extracted from intermediate layers of language models \cite{JAINNEURIPS2018, TONEVANEURIPS2019, schrimpf2021neural, Caucheteux2020.07.03.186288}. Voxel-wise encoding models $\hat{R}$ were estimated using linear regression, and noise covariance $\Sigma$ was computed with a bootstrapping procedure that used the residuals between the encoding model predictions and the actual responses to held-out stories.

The probability that the participant heard a word sequence $S$, $P(S|R)$, can be approximated as $P(R|S) P(S)$. However, it is computationally infeasible to evaluate $P(R|S)$ and $P(S)$ for all possible word sequences. Instead, the most likely word sequence can be approximated by iteratively constructing word sequences using a beam search algorithm \cite{tang2023semantic, tillmann2003word}. In beam search, the $k$ most likely word sequences are retained for each time step. The language model then proposes likely continuations for each sequence on the beam by sampling from $P(S)$ and the encoding model evaluates the probability of the observed brain responses, $P(R|S)$, under each continuation. The $k$ most likely continuations are retained on the beam for the next time step. This process is repeated until the entire scan is decoded.

\subsection{Evaluating Decoding Models}

Decoders were evaluated on brain responses while participants listened to a test story that was not used for model training. Decoding performance was quantified by comparing the decoder predictions for a participant to the actual words that the participant heard \cite{tang2023semantic}. There are many metrics for quantifying the similarity between two word sequences. Because this decoder operates on semantic representations, we quantified similarity using BERTScore, which is a metric designed to quantify similarity of meaning \cite{zhang2019bertscore}. In BERTScore the two word sequences are embedding as semantic vectors using a neural network language model, and a similarity score is computed using the semantic vectors. We computed the BERTScore between the predicted and actual words in every 20-second window of the test story, and averaged the scores across windows to summarize decoding performance for the entire story. These scores were reported as the number of standard deviations above chance level, which was determined by generating word sequences using the language model without using any brain data.

While BERTScore is useful for measuring semantic similarity, the values themselves are somewhat arbitrary and do not correspond to any intuitive notion of decoder quality. To resolve this issue, in addition to the BERTScore metric, we also evaluate our decoding models on an \textit{identification metric} that we refer to in this paper as LogRank. This metric measures how well a given decoder can identify the correct stimulus for a given response among a set of distractor stimuli. Let $E_\theta$ be an estimator of $P(R|S)$ for arbitrary stimulus-response pairs parameterized by our encoding model and noise covariance estimation, jointly described by $\theta$. The LogRank of a particular stimulus-response pair $(S,R)$ is defined with respect to a distractor stimuli dataset $\mathcal{D}$ as 

\[
\text{LogRank}_\mathcal{D}(S, R) = \log_2 \left( \textbf{rank}\left(E_\theta(S, R), E_\theta(\mathcal{D}, R)\right) \right)
\]

 Here, \( \textbf{rank}\left(E_\theta(S, R), E_\theta(\mathcal{D}, R)\right) \) represents the ordinal rank of the estimated probability \( E_\theta(S, R) \) among all estimated probabilities \( E_\theta(S', R) \) for \( S' \in \mathcal{D} \).
 That is, for a given stimulus-response pair $(S,R)$, the LogRank of that pair is the log-base-2 of the rank of the "ground truth" $P(R|S)$ from among all distractors. Alternatively, LogRank can be thought of as the number of additional bits of information that would be required to correctly resolve $S$ as matching $R$ using the given decoder from among the elements of $\mathcal{D}$\footnote{Note that computing this metric does not require that any elements of $\mathcal{D}$ have measured responses.}. For example, if $\mathcal{D}$ contained all possible stimuli, an average LogRank of 0 would correspond to an oracle decoder that is perfectly able to recover the stimulus that corresponds to any response.  This particular interpretation is useful for studying the informatic properties of decoding models and is used frequently in this work. 
 
\subsection{Techniques for Improving Decoding}

We provide and analyze the effects of two methods for improving Bayesian decoding models. These methods are Minimum Bayes Risk (MBR) decoding and encoding model scaling. Each method trades additional computational effort in exchange for improved decoding performance.

\subsubsection{Minimum Bayes Risk Decoding}

Maximum a priori (MAP) decoding tends to fall into pathological local minima when generated sequences deviate significantly from the data that $E_\theta$ was trained on. This effect is especially pronounced with the overlapping mix of semantic information that can be extracted from brain signals. Fortunately, these pathological local minima tend to be transient and highly dependent on the parameterization of $E_\theta$. Our proposed solution to this problem is to augment the standard MAP beam search with a Minimum Bayes Risk (MBR) ensembling approach that averages out pathological examples while retaining semantic themes common across many slightly different versions of $E_\theta$. Let $N$ be the number of ensembled decoding models. We use a leave-one-out approach where we train $N$ separate instances of $E_\theta$ by removing one stimulus from the $E_\theta$ training dataset and retraining on the remaining data. For each of these $N$ leave-one-out generated models, we generate a sequence of words $S_n = \{w_1, w_2, ..., w_l\}$. We then ensemble the set of sequences $\{S_1,S_2,..., S_N\}$ as follows: We instantiate a new beam search with the same parameters, except instead of trying to maximize $P(R|S)$ with our search, we now instead optimize 

$$\max_T\sum_{i=1}^{N} \textbf{BERTScore}(S_i, T)$$

That is, we use beam search to find a target sequence $T$ that is most semantically similar to all other generated sequences according to BERTScore. As a practical consideration, we use a sliding window of the past 20 words when computing BERTScore during decoding to ensure that there are no major temporal deviations from the original sequences. Notably, this approach does not utilize any information from the future to decode the past, so it remains feasible for online decoding. The computational cost of this ensembling scales linearly in the size of $N$. In this work, we test MBR ensembles up to size $N=50$.

\begin{figure}[h!]
    \centering
    \includegraphics[width=\textwidth]{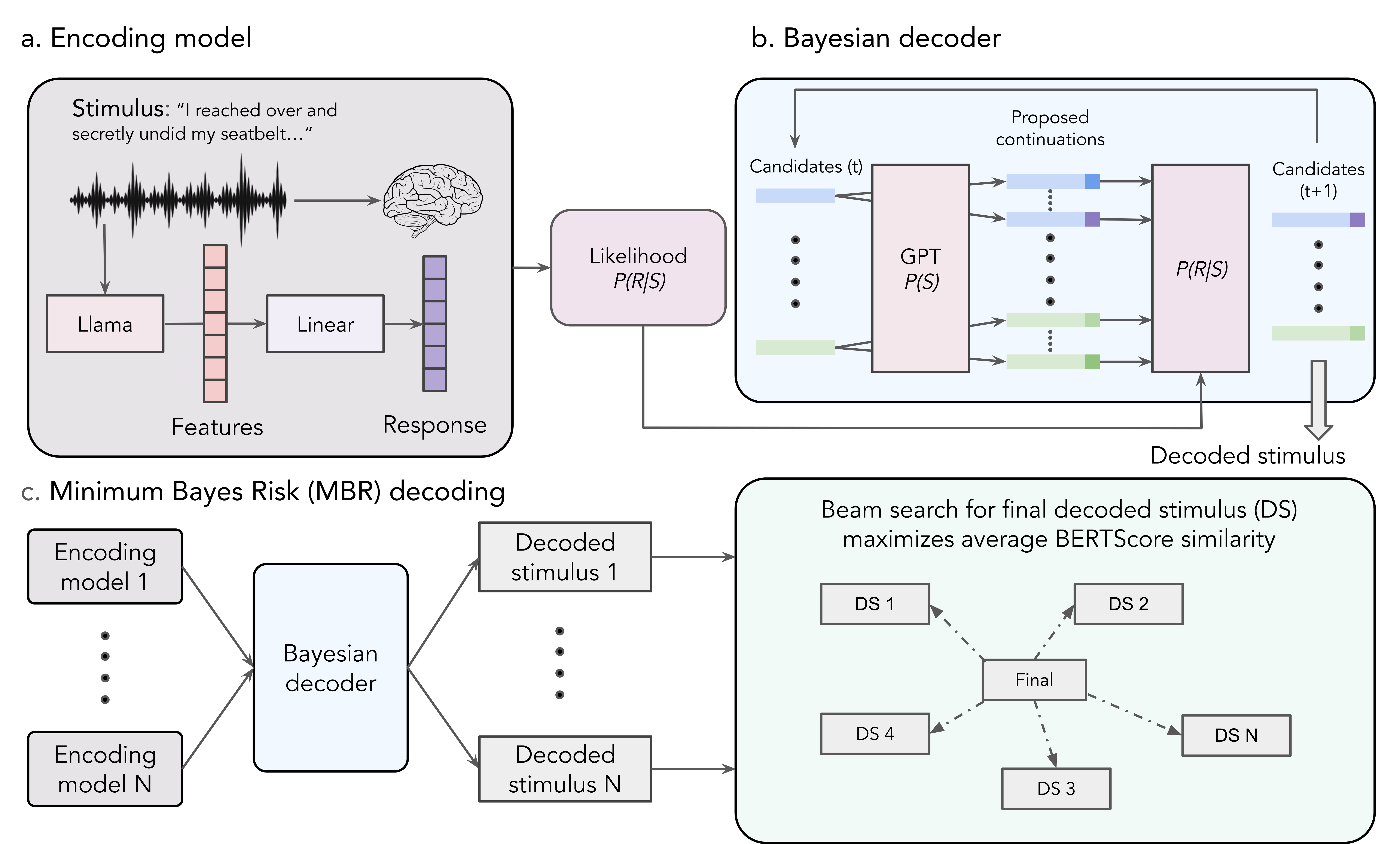}
    \caption{\textit{Decoding methods.} (\textbf{a}) Subjects listen to natural speech while blood-oxygen-level-dependent (BOLD) brain responses are recorded using fMRI. Encoding models use linear regression to predict BOLD responses from features extracted from the stimuli using Llama. (\textbf{b}) A Bayesian decoder uses the encoding model to reconstruct stimulus words from fMRI data. A beam search is performed over word sequences, with candidate beam continuations sampled from GPT-1. The probability of observing the brain responses given the proposed sequences is then evaluated using the encoding model. The best candidate sequences are preserved for the next step. (\textbf{c}) Decoding performance can be improved by ensembling via minimum Bayes risk (MBR). An ensemble of encoding models are estimated by sampling the training data, then are used to decode word sequences. A second beam search then finds a word sequence that is maximally similar to all the ensemble decoded sequences.}
    \label{fig:dummy}
\end{figure}

\subsubsection{Encoding Model Scaling}

Past studies have shown that scaling is an effective means of improving encoding model prediction performance \cite{antonello2023scaling}. We compare a decoding model that uses a finetuned version of GPT-1 (125 million parameters) as a $P(R|S)$ estimator to parameterize $E_\theta$ against one that uses a pretrained Llama-2 model (13 billion parameters). Hidden states were extracted from the 9th layer of the GPT-1 model and the 14th layer of the Llama-2 model. For the optimal decoding analyses in Section 3.2, we used the 14th layer of the pretrained Llama-3 model (8 billion parameters).

We generally follow the method described by Antonello et al. \cite{antonello2023scaling} for encoding model generation of the larger Llama encoding models. For the smaller encoding models, we use the method and code provided Tang et al. \cite{tang2023semantic}. The main practical difference is that the larger encoding models are trained on embeddings extracted from longer context windows. These context window sizes for each model were selected based on coordinate descent. In all cases the same GPT-1 model was used as a language model to estimate $P(S)$. This was done intentionally to mitigate the effects of bias in the LM's training dataset for comparison purposes. To accommodate onset-time-correlated artifacts that manifest in long-context encoding models, a suite of stepwise linear functions of the form $y = \text{floor}(k, t)$ was regressed out of the test fMRI data for those models, for varying values of $k$.

\subsection{MRI data}


We used publicly available functional magnetic resonance imaging (fMRI) data collected from 3 human subjects as they listened to 20 hours of English language podcast stories over Sensimetrics S14 headphones. Stories came from podcasts such as \textit{The Moth Radio Hour}, \textit{Modern Love}, and \textit{The Anthropocene Reviewed}. Each 10-15 minute story was played during a separate scan. Subjects were not asked to make any responses, but simply to listen attentively to the stories. For encoding model training, each subject listened to roughly 95 different stories, giving 20 hours of data across 20 scanning sessions, or a total of   \textasciitilde33,000 datapoints for each voxel across the whole brain. For model testing, the subjects listened to two test stories 5 times each, and one test story 10 times, at a rate of 1 test story per session. The test story with 10 repeats was used for the majority of experiments, whereas all three were used during general informatic analyses (Fig \ref{fig:gpt_llama_comparison}c). Decoding was always performed using the first repeat of a given story, with additional repeats being used for the analysis in Section 3.2. 


Details of the MRI methods can be found in the original publications \cite{lebel2022natural,tang2023semantic,antonello2023scaling}, but important points are summarized here. MRI data were collected on a 3T Siemens Skyra scanner at The University of Texas at Austin Biomedical Imaging Center using a 64-channel Siemens volume coil. Functional scans were collected using a gradient echo EPI sequence with repetition time (TR) = 2.00 s, echo time (TE) = 30.8 ms, flip angle = 71°, multi-band factor (simultaneous multi-slice) = 2, voxel size = 2.6mm x 2.6mm x 2.6mm (slice thickness = 2.6mm), matrix size = 84x84, and field of view = 220 mm. Anatomical data were collected using a T1-weighted multi-echo MP-RAGE sequence with voxel size = 1mm x 1mm x 1mm.

In addition to motion correction and coregistration \cite{lebel2022natural}, low frequency voxel response drift was identified using a 2nd order Savitzky-Golay filter with a 120 second window and then subtracted from the signal. The mean response for each voxel was subtracted and the remaining response was scaled to have unit variance. 

The original publications on this dataset stated that all subjects were healthy and had normal hearing. The experimental protocol was approved by the Institutional Review Board at The University of Texas at Austin. Written informed consent was obtained from all subjects.

\begin{table}[h]\centering
    \centering
    \begin{tabular}{lccccccccccccc}
         & \multicolumn{1}{c}{GPT-1} & \multicolumn{1}{c}{Llama-2} & \multicolumn{3}{c}{GPT-1 + MBR} & \multicolumn{4}{c}{Llama-2 + MBR} \\
        \cmidrule(r){2-2} \cmidrule(r){3-3} \cmidrule(r){4-6} \cmidrule(r){7-10}
        Subj.& Single & Single & 5 runs & 10 runs & 20 runs & 5 runs & 10 runs & 20 runs & 50 runs \\
        \midrule
        S01 & 14.6 & 16.7 & 16.8 & 16.8 & 17.6 & 19.5 & 21.2 & \textbf{22.1} & 21.4 \\
        S02 & 14.6 & 15.9 & 17.2 & 16.7 & 17.8 & 18.2 & 19.7 & 21.0 & \textbf{21.4} \\
        S03 & 18.7 & 20.0 & 21.9 & 22.4 & 23.5 & 21.9 & 23.4 & 25.4 & \textbf{26.1} \\
        Avg & 16.0 & 17.5 & 18.6 & 18.6 & 19.6 & 19.8 & 21.4 & 22.8 & \textbf{23.0} \\
    \end{tabular}
    \linebreak
    \caption{Ablation results showing improvements in BERTScore for each of the methods as well as the sum of both methods. BERTScore is measured in standard deviations from a random baseline generated by the GPT-1 language model without consideration of $P(R|S)$). For
    all subjects, we see improvements from 40-50\%.}
    \label{tab:ablation}
\end{table}
\subsection{Estimating an idealized semantic decoding model}

The encoding model predictions for a stimulus should ideally match the participant's actual brain responses to that stimulus. In practice, encoding models may differ from actual brain responses, leading to suboptimal decoding performance.

One potential source of decoding error is encoding model misspecification, such as extracting suboptimal features from language stimuli. To characterize decoding performance using an optimal encoding model, we performed identification for a test story using averaged responses across 9 subsequent repeats of that story. The averaged responses provide a noise ceiling that captures the explainable signal in each voxel. The difference in identification performance between the encoding model and the inter-trial ceiling quantifies the amount of improvement that can be attained by using a better encoding model.

Another potential source of decoding error is measurement noise. If there is structure to these errors across voxels, it can bias decoder predictions. As a result, it is useful to account for this structure when comparing the encoding model predictions and the actual recorded responses \cite{van2018modeling}. Previous studies have modeled the structure of the noise by computing the covariance of the encoding model residuals on held out stories not used in encoding model training \cite{tang2023semantic, nishimoto2011reconstructing}. To assess whether improving the noise model will improve decoding performance, we compared identification performance for the actual noise model with identification performance for a ceiling noise model. In order to isolate the structure of noise that is repeatable across trials, we used brain responses while the participants listened to 9 subsequent repeats of the test story. We computed the true noise covariance for each of the repeats, and averaged covariance matrices across repeats to create a ceiling noise model. The difference in identification performance between the estimated noise model and the inter-trial noise model quantifies the amount of improvement that can be attained by using a better noise model.

For each idealized model, we evaluate the number of bits that have been extracted from the model using the LogRank metric. We perform a ``Pareto correction'' on the data, which helps us resolve instances when $|\mathcal{D}|$, the number of distractors tested, is not large enough that any distractor stimulus is preferred to the true stimulus. This causes a significant fraction of our data to be ``censored'', meaning we cannot resolve the true number of bits beyond a certain threshold, in this case, beyond $\log_2 |\mathcal{D}| - 1$. In these cases, we fit a Pareto distribution to the existing data, and extract a power-law relationship $y=kx^a$. This relationship is then interpolated past the resolution threshold. Since the area under a power-law is undefined, the censored probability mass is distributed as conservatively as possible below the extrapolated relationship. Let $p_{\text{censored}}$ be the probability mass of the Pareto distribution that was censored. For each distribution, we compute the value of $x_{\text{min}}$ such that
\[\int_{x_{\text{min}}}^\frac{1}{|\mathcal{D}|} kx^a \, dx = p_{\text{censored}}\]
and then distribute the remainder of the probability mass between $x_{min}$ and $\frac{1}{|\mathcal{D}|}$. Due to the conservative nature of this estimation procedure, estimates of ``bits extracted'' elsewhere in this paper should be considered lower bounds. 


\subsection{Compute specifications}

The generation and use of the models presented in this paper required significant computational resources. Ridge regression was performed using compute nodes with 128 cores (2 AMD EPYC 7763 64-core processors) and 256GB of RAM. In total, roughly 2,000 node-hours of compute was expended for these models. Feature extraction for language models was performed on specialized GPU nodes similar to the AMD compute nodes but with 3 NVIDIA A100 40GB cards. Feature extraction and subsequent decoding required roughly 4000 node-hours of compute on these GPU nodes. In total, this research used an estimated $5 \times 10^{21}$ floating point operations.

\section{Results}

\begin{figure}
  \centering
  \vspace{-1.5em}
  \includegraphics[width=\textwidth]{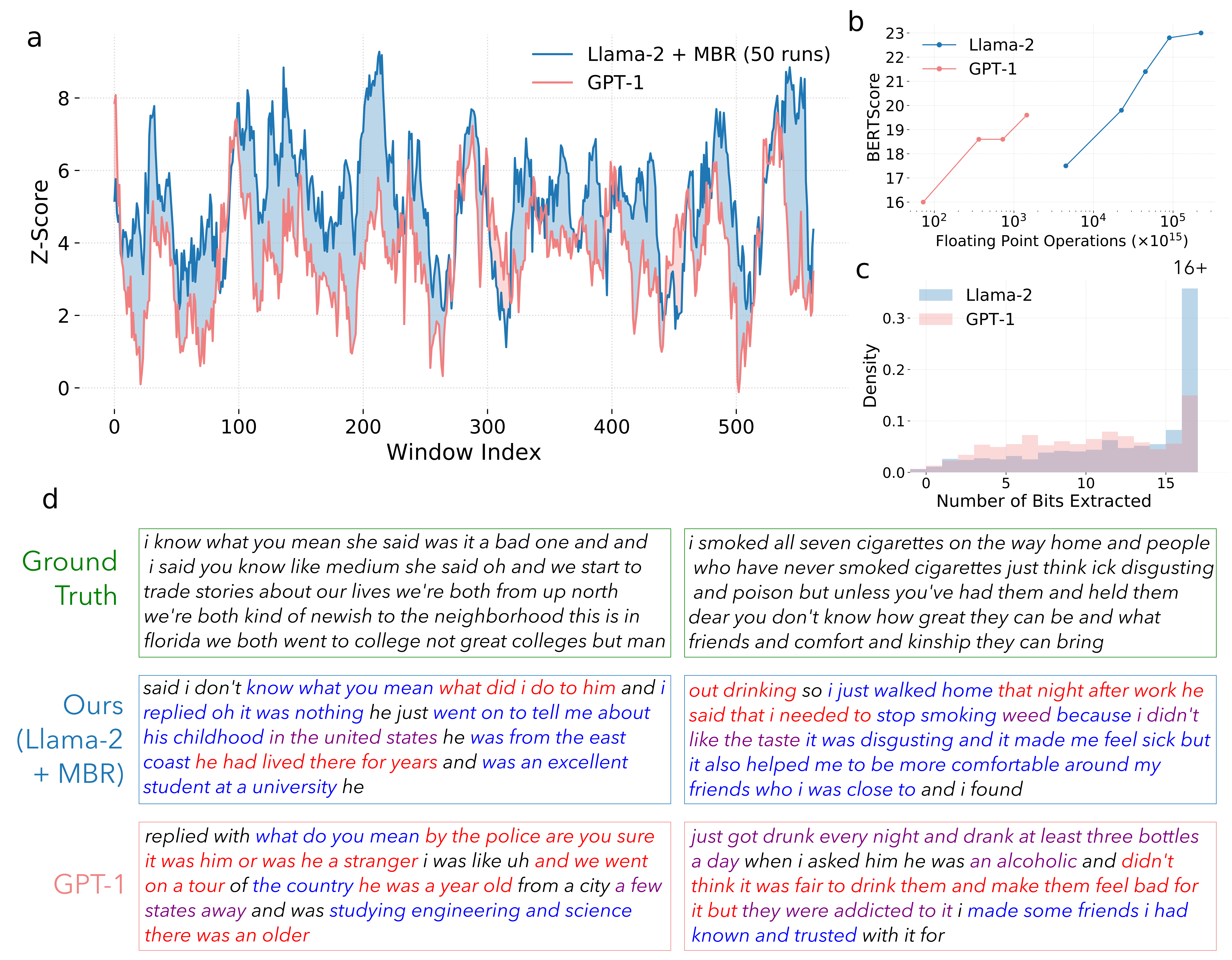}
  \caption{\textit{Comparison of Decoding Models:} (\textbf{a}) \textit{Comparison of Llama-2 with MBR over GPT-1 Baseline}: Semantic similarity of each model output with the ground truth is plotted on the y-axis over the timecourse of a held-out test story. The Llama-2 model with MBR is shown to outperform GPT-1 in almost all cases. (\textbf{b}) \textit{Computational Costs of Decoding}: A log-plot comparing computational cost aginst estimated performance. A clear cost-performance tradeoff is visualized. (\textbf{c}) \textit{Number of Bits Extracted:} The estimated amount of information extracted using the $P(R|S)$ approxmiators built from each model is visualized. Llama-2 extracts about 2.8 more bits of information, after Pareto correction. The number of bits is estimated by the number of distractors that are ranked better than the ground truth. (\textbf{d}) \textit{Qualitative Performance Comparison}: Two well-decoded extended contexts from each decoding model (Llama-2 with a 50 run MBR ensemble, GPT-1) are presented. The ground truth text is presented bordered by green. For each decoding, words are colored based on whether they are semantically-correct (\textit{blue}), gist-capturing (\textit{purple}), or incorrect (\textit{red})}.
\label{fig:gpt_llama_comparison}
  \vspace{-1.5em}
\end{figure}

\subsection{Improving Decoding Performance}

\textbf{Table \ref{tab:ablation}} demonstrates the main performance improvements of our brain-to-text decoding methods. Across all subjects, MBR with Llama-2 results in an estimated 40-50\% improvement in measured BERTScore similarity to the ground truth story, with the average BERTScore across subjects increasing from 16.0 to 23.0. Examining the ablation results, we see that the majority of the contribution comes from MBR rather than scaling. However it also appears that there are some second-order effects, with scaling having greater power when combined with MBR than when used without ensembling. In particular, the benefit of 20 runs of MBR is only 2.1$\sigma$ for GPT-1 on average, whereas it is 5.3$\sigma$ for Llama-2. This suggests that the implicit prior adopted by beam search may be ill-suited to the problem of continuous brain-to-text decoding, and that the semantic ensembling procedure ameliorates this problem somewhat, with greater benefits for better models. This result also points to an underexplored avenue for future decoding research, as it may be the case that Bayesian decoding models could benefit more from adoption of more effective priors, rather than simply better estimation of $P(R|S)$ and $\Sigma$.

\textbf{Figure \ref{fig:gpt_llama_comparison}a} illustrates the performance improvement across the timecourse of a test story. A sliding window of the BERTScore of each method's output against the ground truth is plotted. We see that, for nearly every TR, the fully ensembled (50 run) Llama-2 decoding model outperforms the GPT-1 decoder. The average window score rises here from 3.7$\sigma$ for the GPT-1 encoder to 5.2$\sigma$ for the Llama-2 decoder, over a 40\% increase.

\textbf{Figure \ref{fig:gpt_llama_comparison}b} shows the relationship between decoding performance and inference costs, with higher costs associated with better performance. We note that the Pareto frontier suggested by this plot is to first ensemble and then only when reaching diminishing returns, use scaling as a means to further improve performance. As the 20-run GPT-1 ensemble is both cheaper and performs better than the single-run Llama model,  it is probable that there are further improvements to be made on the algorithmic side in terms of computational efficiency and more effectively managing this tradeoff. We note that the current computational cost of our best decoder is largely impractical for virtually all non-research settings, requiring about 500 hours of A100 runtime to decode a 15 minute fMRI scan. Further algorithmic work into reducing this cost to enable online decoding is essential if these methods are to have practical value.

\textbf{Figure \ref{fig:gpt_llama_comparison}c} compares the number of bits extracted by each $P(R|S)$ encoding model estimator on subject S3. We see that the Llama-2 encoding model is consistently able to outperform the GPT-1 encoding model, with over 30\% of all \textasciitilde900 test TRs being correctly identified from among over 140,000 distractor stimuli. After performing the Pareto correction described in Section 2.5, we estimate that the GPT-1 decoder extracts 11,1 bits on average for a given TR, whereas the Llama-2 decoder extracts 13.9 bits, a 2.8 bit correction. Versions of \textbf{Figure \ref{fig:gpt_llama_comparison}c} for other subjects are provided in the Appendix. 

\textbf{Figure \ref{fig:gpt_llama_comparison}d} provides some qualitative examples of decoder quality comparing the two models on well-decoded TRs. The Llama-2 model (\textit{blue}) is able to sometimes capture complex and nuanced sentiments, in comparison to the GPT-1 model, which often correctly captures simple semantic themes but does not successfully tie them together. Full decoded texts for this test story using the 50-run MBR ensemble are provided in the Appendix for each of the three test subjects.

\subsection{Estimating Idealized Decoding Models}

\begin{figure}
  \centering
  \vspace{-1.5em}
  \includegraphics[width=\textwidth]{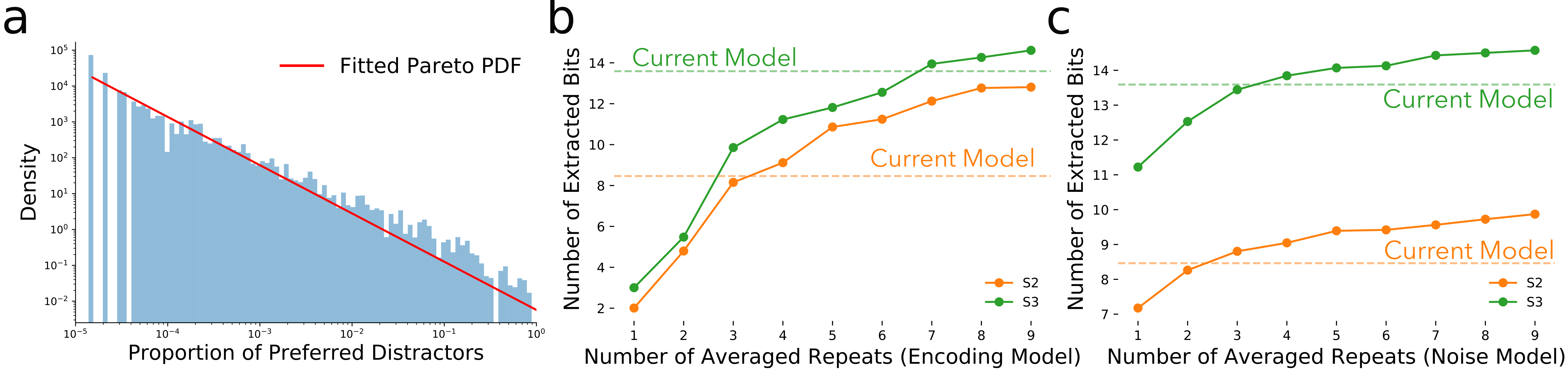}
  \caption{\textit{Informatic properties of Bayesian decoding} (\textbf{a}) \textit{Power law dynamics}: Plotted is the proportion of distractors from the total set that are evaluated as having a higher $P(R|S)$ over the ground truth stimuli for S3. The relationship obeys a power law and can be modelled with a Pareto distribution (\textit{red line}). (\textbf{b}) \textit{Idealized encoding performance}:  Idealized encoding models were computed by averaging responses across different repeats of the test story. With 9 repeats of the test story, identification using averaged responses led to an average 2.7 bits of improvement over the current state-of-the-art encoding model. (\textbf{c}) \textit{Idealized noise estimation}: Idealized noise models were computed by averaging noise covariance matrices across different repeats of the test story. With 9 repeats of the test story, identification using idealized noise models led to an average 1.2 bits of improvement over the current state-of-the-art noise model. S1 is omitted from these analyses due to poor test story repeatability. These results suggest further room for improvement in $P(R|S)$ estimation.}
  \vspace{-1.5em}
\label{fig:idealized_em_nm}
\end{figure}

In addition to improving the performance of these models, we also studied their statistical and informatic properties. \textbf{Figure \ref{fig:idealized_em_nm}a} plots a log-log histogram showing the distribution of the proportion of preferred distractors to total distractors for each of our \textasciitilde900 test TRs. This number includes the ground truth stimulus itself, so the numerator is always at least 1. A lower number indicates that very few distractors are rated as more likely to cause the observed fMRI data than the true stimulus, with a proportion of $\frac{1}{|\mathcal{D}|}$ corresponding to ideal performance. We observe that the relationship is governed by a Zipfian power law and can be therefore modelled by a Pareto distribution \cite{newman2005power}. As word frequencies are themselves governed by a similar distribution, it may be reasonable to assume that this relationship is a byproduct of the statistics of natural language itself rather than of the biological and physical processes in the brain.

\textbf{Figure \ref{fig:idealized_em_nm}b} explores how many bits of information could additionally be extracted over our current model from brain responses if our encoding model $E_\theta$ was optimal. To get an estimate of the noise ceiling, we averaged the responses across several repeats of the same test story. We then substituted the predicted response in these for these averaged responses and then computed the increase in the average number of extracted bits. We find that, with an optimal encoding performance, we would be able to extract an additional 2.7 bits of information over our current state-of-the-art models. Due to the conservative nature of the "bits extracted metric (see Section 2.5), this is probably a lower bound, with additional room for improvement likely. 

\textbf{Figure \ref{fig:idealized_em_nm}c} answers a similar question, but from the perspective of improving our noise covariance estimation. We examined how many additional bits of information could be extracted from our current model with better covariance estimatation procedures. To do this, we averaged the noise covariance across repeats of the same test story and substituted this averaged covariance as a replacement for the existing noise covariance estimation. We find that better noise covariance estimation could yield at least an average additional 1.2 bits of improvement over the current model.

\section{Discussion}

In this paper, we have demonstrated a substantial improvement in the state-of-the-art for non-invasive semantic brain-to-text decoding, improving existing models by upwards of 40\% on our held-out test stimulus. The improvement relies on two primary modifications. First, we found that decoding performance can be improved by using more powerful language models during encoding, reflecting previous findings that encoding model performance scales with the effectiveness of the language model used to extract features \cite{antonello2023scaling}. If recent trends in language model scaling continue, we expect that decoding performance will continue to improve as well. Second, we show that decoding performance can also be improved by developing new algorithms that leverage the growing amount of computing power through methods such as ensembling. These techniques may provide an avenue for improving decoding performance orthogonal to improvements in language models. 

In addition to these significant empirical improvements, our study also characterizes the gap between current language decoding performance and the best possible performance that can be achieved using fMRI. We found that using an ideal encoding model could increase identification performance by an average of 2.7 bits, while using an ideal noise model could improve identification performance by an average of 1.2 bits. These estimates are only lower bounds, and indicate that there is a substantial amount of improvement that can be attained purely through algorithmic advances. Further enhancements may also arise from developing more effective methods for translating the information extracted by these models into natural language.

Better decoding performance has critical practical implications. Currently, motor decoders that use invasive brain recordings are sufficiently accurate to improve communication for people who have lost the ability to speak due to injuries such as strokes, or diseases such as amyotrophic lateral sclerosis \cite{willett2023high, metzger2023high}. Our results show that non-invasive semantic decoding could eventually provide a safer and cheaper alternative to these ECoG-based decoding methods that require costly neurosurgery.

On the other hand, improved decoding performance has important implications for mental privacy. Previous studies have found that semantic decoders only produce accurate predictions when brain data are collected with a participant's cooperation \cite{tang2023semantic}. For instance, participants could actively prevent decoders from predicting a story that they are hearing just by thinking about something else. It is important to continually evaluate these models as decoding algorithms become more powerful. 

The relative ease with which we were able to substantially boost the state-of-the-art for semantic decoding suggests that additional advancements are likely to follow in the near future. Society should prepare for this outcome by adopting forward-looking mental privacy standards and regulations. 

\section*{Acknowledgements}

The authors acknowledge and thank the Texas Advanced Computing Center (TACC) at The University of Texas at Austin for providing HPC resources that have significantly contributed to the research
results reported within this paper. This work was funded by grants from the NIDCD and NSF (1R01DC020088-001), the Burroughs-Wellcome Foundation, a gift from Intel Inc, and the support of the William Orr Dingwall Foundation. We thank
Aditya Vaidya and Matthew Weidner for their aid and thoughtful suggestions in assisting with
this work.

\footnotesize
\bibliographystyle{unsrtnat}
\bibliography{bibliography.bib}


\clearpage

\appendix

\setcounter{table}{0}
\setcounter{figure}{0}

\renewcommand{\thetable}{A\arabic{table}}
\renewcommand*{\thefigure}{A\arabic{figure}}

\section{Appendix / supplemental material}

\subsection{GPT-1 vs. Llama-2 comparisons (other subjects)}

Here we show copies of \textbf{Figure \ref{fig:gpt_llama_comparison}a} and \textbf{Figure \ref{fig:gpt_llama_comparison}c} using the data from other subjects, as the figure in the main text was generated with S03. 

\begin{figure}[h]
  \centering
  \includegraphics[width=\textwidth]{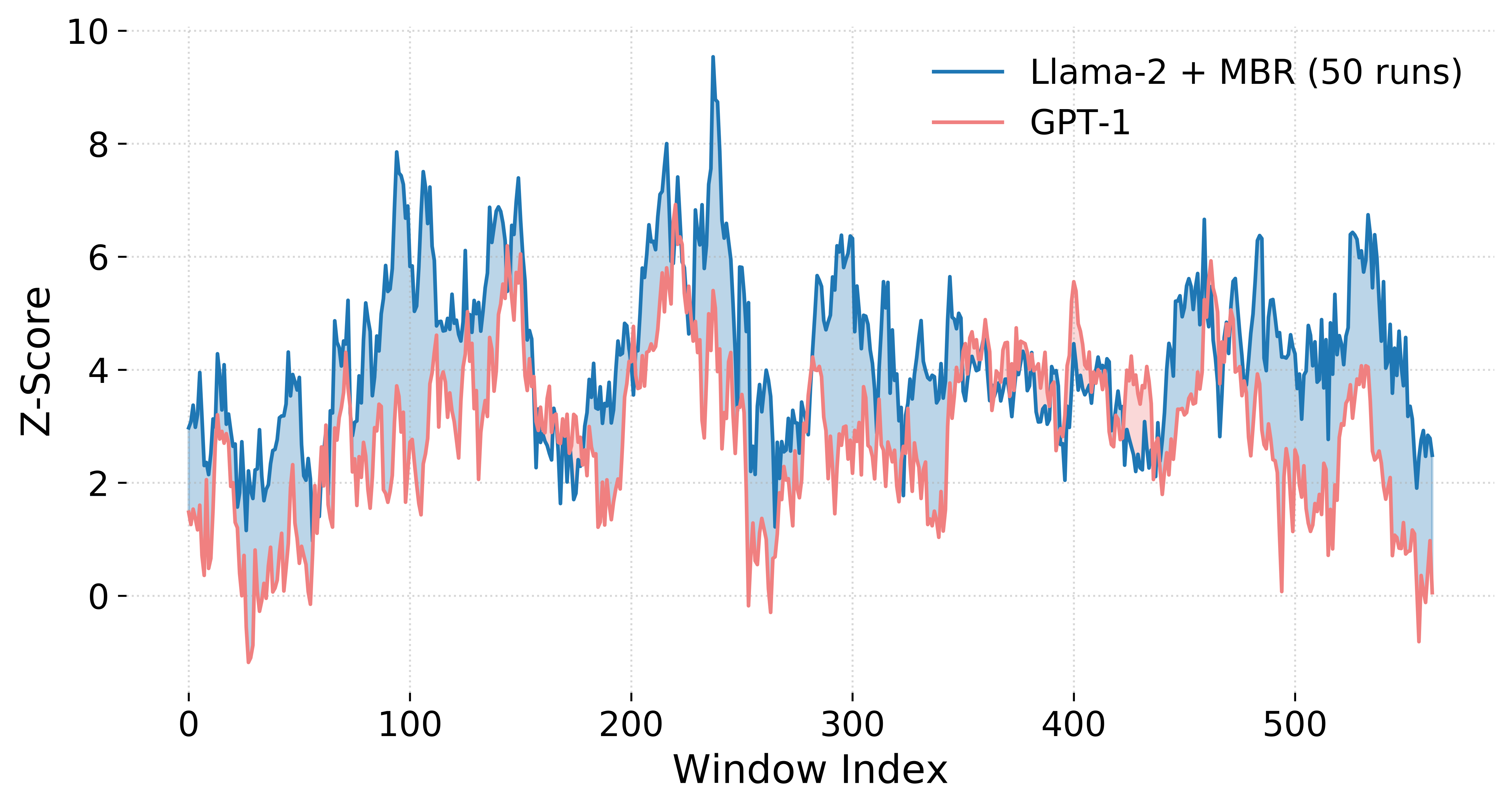}
  \caption{GPT-1 vs. Llama-2 + MBR (50 run) comparison for S01}
\label{fig:comparison1}
\end{figure}

\begin{figure}[h]
  \centering
  \includegraphics[width=\textwidth]{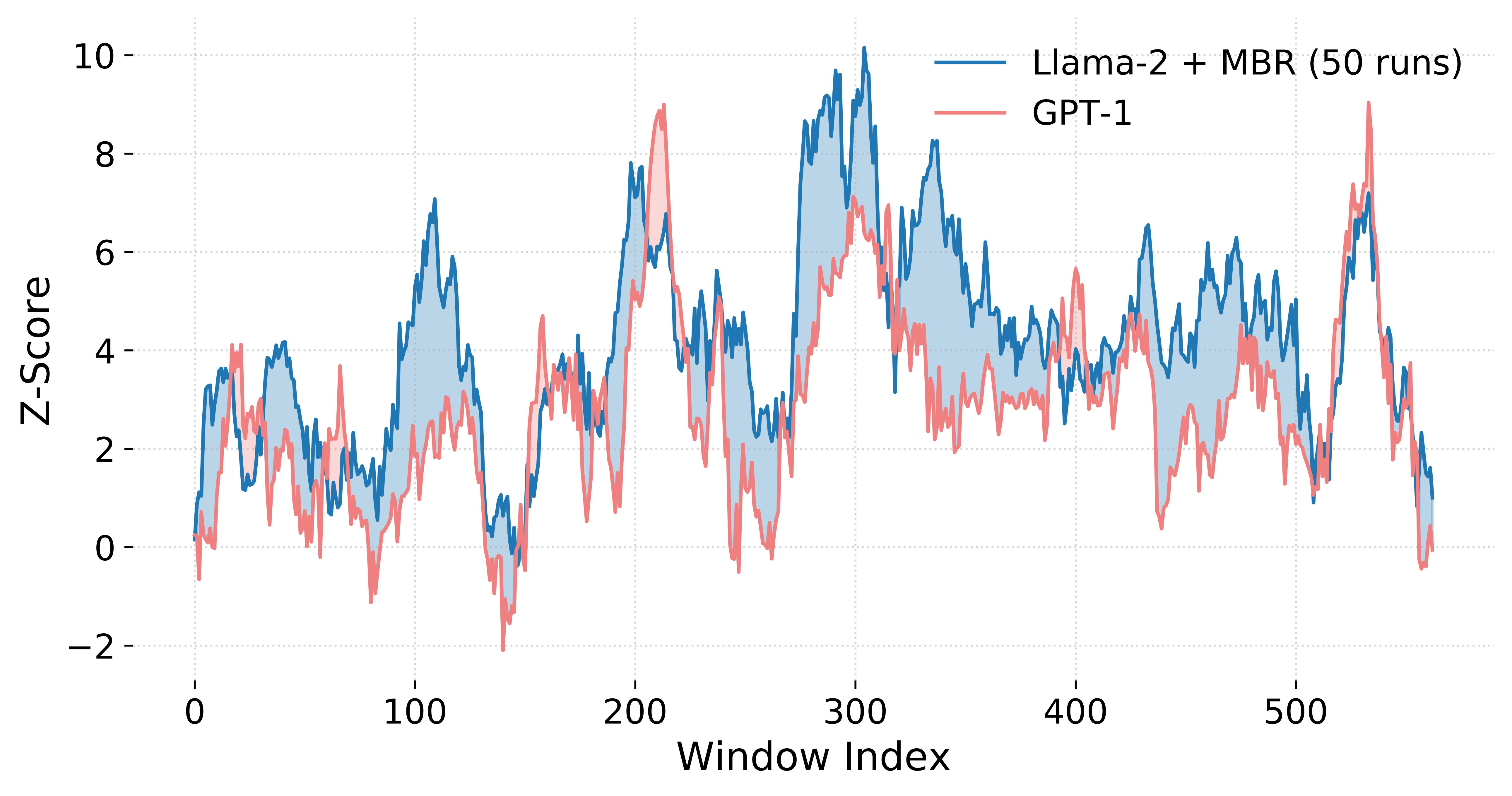}
  \caption{GPT-1 vs. Llama-2 + MBR (50 run) comparison for S02}
\label{fig:comparison2}
\end{figure}

\clearpage

\begin{figure}[h]
  \centering
  \includegraphics[width=\textwidth]{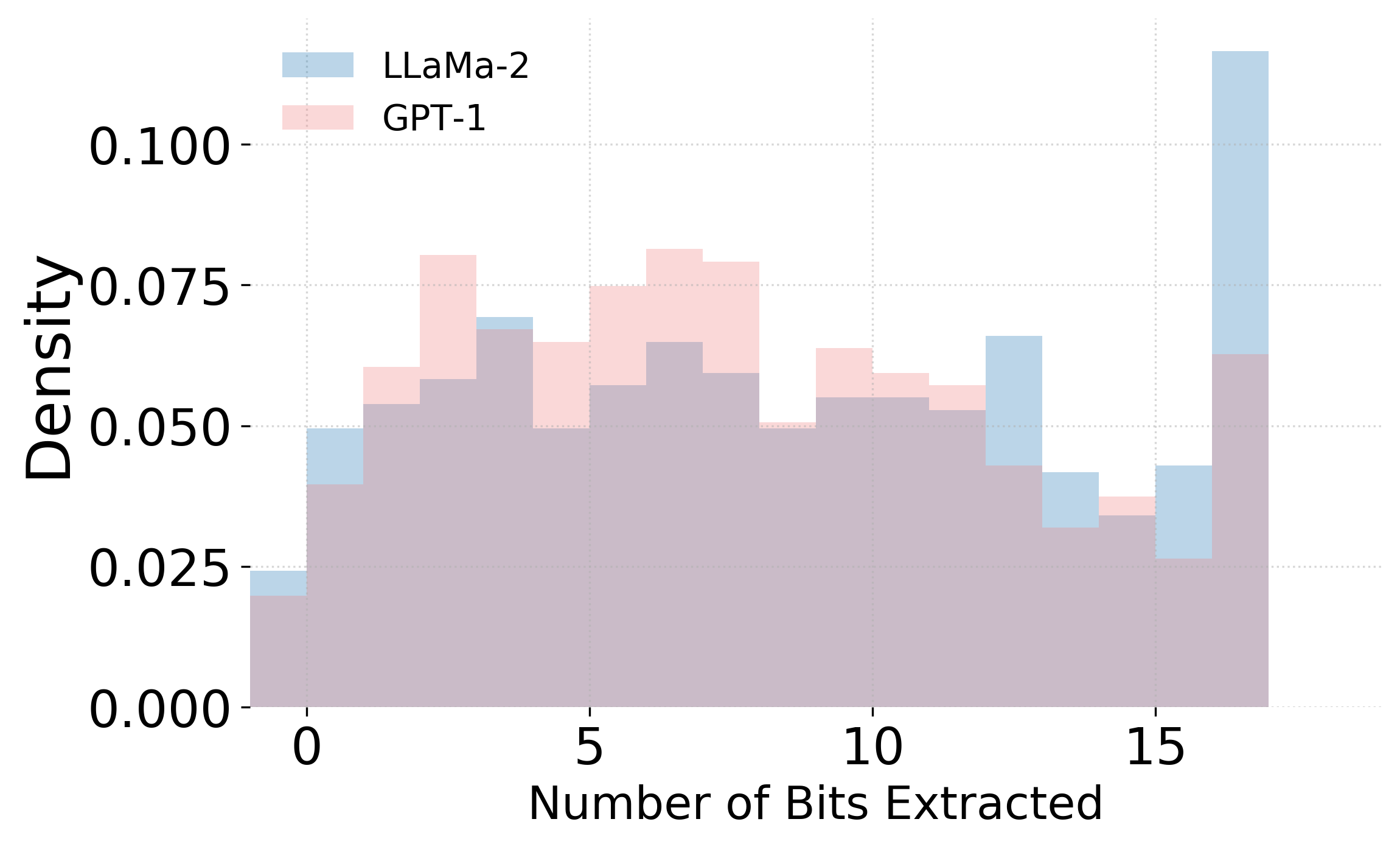}
  \caption{Bits extracted for S01}
\label{fig:bits1}
\end{figure}

\begin{figure}[h]
  \centering
  \includegraphics[width=\textwidth]{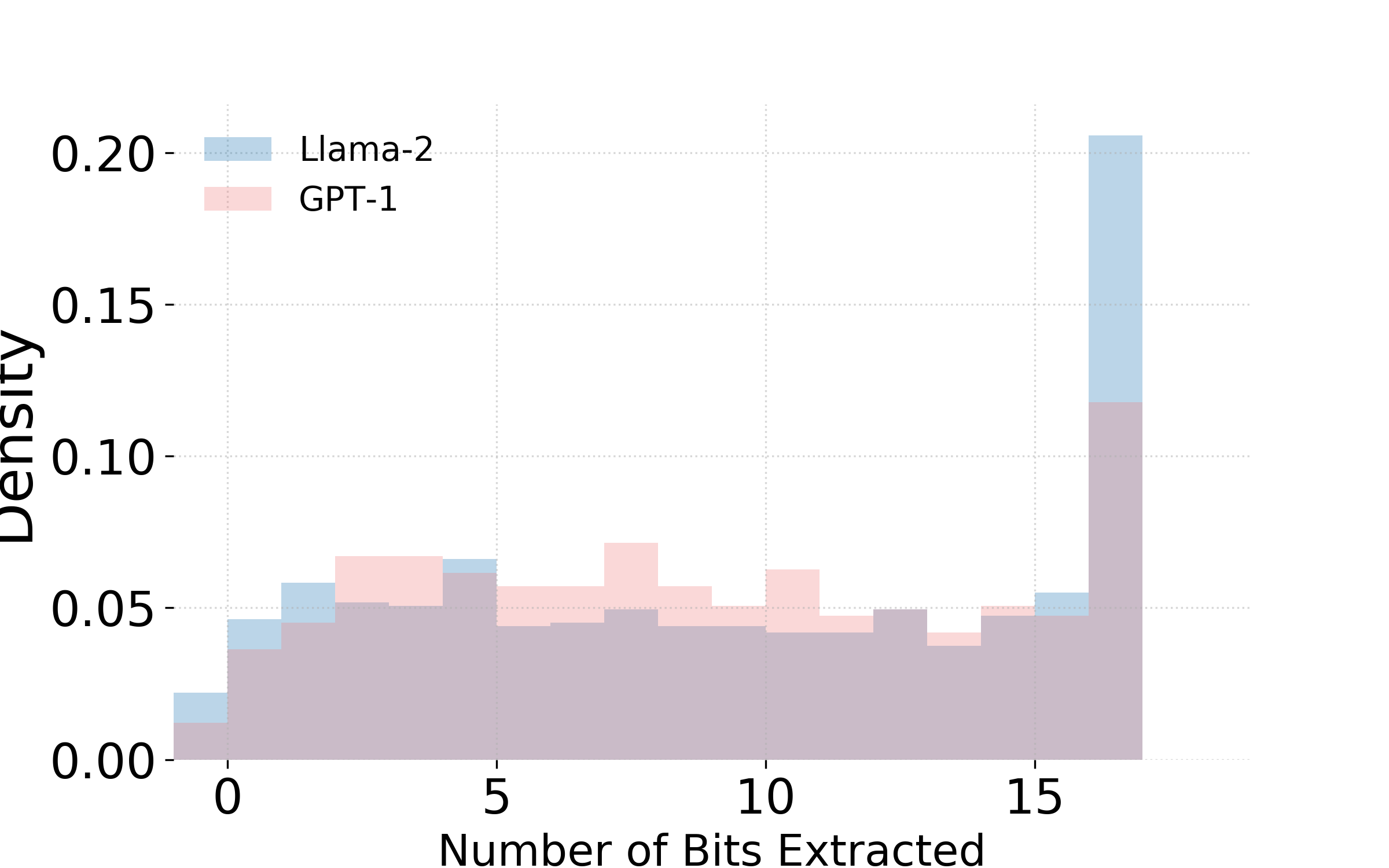}
  \caption{Bits extracted for S02}
\label{fig:bits2}
\end{figure}

\clearpage

\subsection{Power law dynamics (other subjects)}
Here we show copies of \textbf{Figure \ref{fig:idealized_em_nm}a} generated using the data from other subjects, as the figure in the main text was generated with S03. 
\begin{figure}[h]
  \centering
  \includegraphics[width=\textwidth]{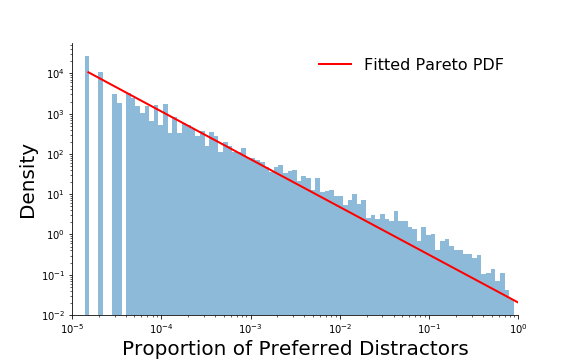}
  \caption{Power law dynamics of proportion of preferred distractors for S01}
\label{fig:pareto1}
\end{figure}

\begin{figure}[h]
  \centering
  \includegraphics[width=\textwidth]{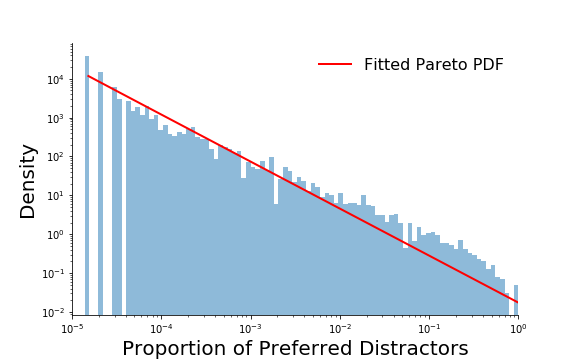}
  \caption{Power law dynamics of proportion of preferred distractors for S02}
\label{fig:pareto2}
\end{figure}

\clearpage

\subsection{Full text of Llama2 + 50-run MBR ensembled decoded texts}

\hspace{-3em}
\begin{tabular}{|c|c|c|c|}
    \hline
    \textbf{Ground Truth} & \textbf{S01} & \textbf{S02} & \textbf{S03} \\
    \hline
    i had no shoes on & long day i didn't& only reason people have&job i have no\\
    i was crying i had no wallet & sleep much i was&trouble finding jobs i &savings but i'm not \\
    but i was ok because & exhausted but i could & know because i do it&homeless i can afford\\
    i had my cigarettes &have done with a cup &  every day i don't &to buy groceries and\\ 
    and i didn't want &of coffee i don't know &mind if you don't &gas i don't need a\\
    any part of freedom & what made me decide&because it doesn't mean &job anymore because i \\ 
    if i didn't have & to go home i guess& you aren't good at&have no money to pay\\
    my cigarettes & the worst thing about &it it's just a natural &for college my dad\\ 
    when you live with & it was that the girl& reaction to things that&is an alcoholic he\\
    someone who has &was really upset and & happened before you know&was always abusive and \\ 
    a temper a very &i was so angry and & what&it was very\\
    bad temper a &  &  &  \\
    \hline
    very very bad temper &sad that i couldn't & to do if you do&difficult for me to \\
    you learn to play &stand to talk to her & something wrong it's&deal with him i was\\ 
    around that you &about it because i & very hard to say it&the one who would\\
    learn this time i'll &didn't know how to & is what you want&say what i needed to\\
    play possum and next & fix it she told me& but you will do&and not what he said \\
    time i'll just be & to just say whatever&what is right in &or if he did so\\
    real nice or i'll & she wanted and that& order to be successful&mething wrong then i\\
    say yes to everything  & i wasn't allowed to&you may not realize &could either argue or just\\
    or you make yoursel &change anything and she & this but your goal is&walk away but that\\
    scarce or you run & didn't care what i&to help your son &doesn't work for me\\
    and this was one &did but i still & succeed but if he&the other reason i did\\
    of the times when you & did it it was hard&fails that means you &this is that\\
    just run and as & and it hurt because i & have to do it alone i &  \\
    
    \hline
    i was running i thought & knew that she had& would say that you&i had to move to\\
    this was a great &the power to do it & should have done this& an area where i\\
    place to jump out & again i went to the& in your home town but&couldn't see anything\\
    because there were big &hospital that day and &this is where most &because the roads were\\
    lawns and there & when they asked me to&of the people live the &too narrow for cars to\\
    were cul de sacs and  & sit down she said that&  police will be there&pass by me at night my\\
    sometimes he would &i should just leave & if they see you walking&dad would just yell\\
    come after me and &her alone and go home & down the street so&at me to go home but\\
    drive and yell stuff &i was like ok no & i say i want to leave& i kept saying no\\
    at me to get back  &thanks and walked out & here so we head&and running off into\\
    in get back in and  &to the parking lot &over to a park a block &the woods he was like\\
    i was like no i'm out  & with my friend we went& from my school and he&hey you can get\\
    of here this is great &back inside the store & gets his phone and &the car back and i\\
    and i went and hid  &and i had my first & starts talking to some of& couldn't believe it\\
    behind a cabana and & drink& &but he had been driving\\
    he left and i had my & & & \\
    cigarettes and uh i & & &\\
    started to walk in this &  &  &  \\
    \hline
    beautiful neighborhood & with him it was about& the students i saw&around town for about\\
    it was ten thirty at &minutes into the show  & on the street and & hours i went to see\\
    night and it was silent & and he was drunk i& it seems that the&him at about pm\\
    and lovely and there & was at a bar with&guy was really good &and he was pretty drunk\\
    was no sound except &  two of my friends& at it and i think he&and the place was\\
    for sprinklers ch ch &and this guy had a &was the one who  &packed and we had a\\
    ch ch {ig} ch ch ch ch &giant beer bottle on &decided to teach the &few beers and smoked\\
    {ig} and i was enjoying  &my bed and i &class a couple days &some weed the whole\\
    myself &  &  &  \\
    \hline
\end{tabular}

\hspace{-3em}
\begin{tabular}{|c|c|c|c|}
    \hline
    \textbf{Ground Truth} & \textbf{S01} & \textbf{S02} & \textbf{S03} \\
    \hline
    and enjoying the absence & was so happy to& after it started but&night and it was \\
    of anger and enjoying &finally be alone &i couldn't focus on &really hard to feel\\
    these few hours i knew & and not have to talk&anything but how it &happy after a few\\
    i'd have of freedom and &about what had happened & affected my life so&months of trying to be\\
    just to perfect it i &in the past few days & much and how i could&the best i could i\\
    thought i'll have a smoke &and how i felt i &not deal with it because& had no money to buy\\
    and then it occurred to me & also started to wonder& i didn't want to feel&anything but i knew it\\
    with horrifying speed i &  if it was something i&that way anymore it &would be impossible to\\
    don't have a light just & did that caused me to& is hard to understand&get a car since i didn't\\
    then as if in answer i & lose it and then i& how it happened i can't&have insurance i asked \\
    see a figure up ahead & went out to the & really explain why but i &  a guy if he was \\
    &street and saw a man&felt the same way when&a cop and\\
    &coming towards&i saw it in the mirror the&\\
    \hline
    who is that it's not &me i thought hey  & night before i thought&he said no but\\
    him ok they don't have & that guy looks like&maybe my parents were &i didn't see any\\
    a dog who is that what &my dad but i don't & in another room but &reason why he would\\
    uh what are they doing & know him well he&i knew they were &be in the store\\
    out on this suburban street &was walking down the &both there it was dark &because of what happened\\
    and the person comes closer &street at this point &out but the hall  &the guy walked in\\
    and i could see it's a woman & and we saw him&light was on and i & and i saw that he\\
    and then i can see she & coming towards us he&could see him moving &had his hand up\\
    has her hands in her face & had his hand up& around he had his&the front of his shirt\\
    oh she's crying and then & to his face and& hand on the wall and& and he was rubbing his\\
    she sees me and she composes & was looking in our&then he was looking &face as i approached he\\
    herself and she gets closer &direction and i was frozen &up at me he then &then turned around and\\
    and i see she has no & in fear and he started & turned and walked away & looked at me and started \\
    &talking to us we were&i ran over to him&crying as he walked back\\
    &&and i saw that&to his car\\
    
    \hline
    shoes on she has no shoes & both staring at him& his face was pale&his head was still\\
    on and she's crying and & he was the same&he had a very &down and i was\\
    she's out on the street & age and his eyes& similar look to the&very scared and confused\\
    street i recognize her &were red from crying & guy i had seen&i think it was\\
    though i've never met her &i remember seeing him  &before but we didn't &because he didn't\\
    and just as she passes &once but didn't know &really talk to him &recognize me but i\\
    me she says you got a & who he was he looked& he came over and he&didn't really know him \\
    cigarette and i say you &at me and asked &asked if we had &then he asked me\\
    got a light and she says & me if i had a& any beer in the car&if i had a girlfriend\\
    damn i hope so and then & girlfriend i said no& he didn't have any&i said no and that\\
    sh first she digs into & but he didn't know& and we said no&i didn't think so\\
    her cutoffs in the & so i went to her & he was a great guy & he did a quick \\
    &room and found a box of&but i didn't&check of my backpack\\
    \hline
    front nothing and then & her clothes i opened& know that until a&and found some paper\\
    digs in the back and then &it up and it & year later when he&towels in the back\\
    she has this vest on that &was like a little & went to jail and&pocket of the pants\\
    has fifty million little &  black bag that had&we all saw the &and some other things\\
    pockets on it and she's & a few things inside&  pictures and the whole&that i wasn't sure\\
    checking and checking and &of it that i couldn't &thing was a mess &where to find but\\
    it's looking bad it's &figure out at first &but i didn't really &they were still there\\
    looking very bad she digs &  and then i just&  know that much about it&i looked at them a\\
    back in the front again & grabbed a small box&until i came to  &bit more and found\\
    deep deep and she pulls & and opened it and& see the house and&a small bag of weed\\
    out a pack of matches that & pulled out a huge& it was a big place&in there that had\\
    had been laundered at & pile of bills i & and everything was nice&probably been around years\\
    least once ukgh we open & had thrown away but it & and all that but & old i didn't want \\
    &was still worth it&i still&\\
    \hline    
\end{tabular}

\hspace{-3em}
\begin{tabular}{|c|c|c|c|}
    \hline
    \textbf{Ground Truth} & \textbf{S01} & \textbf{S02} & \textbf{S03} \\
    \hline
    it up and there is one &so i took it  &got in trouble for &to smoke it so\\
    match inside ok oh my & all home with me &smoking weed after a &i took the joint\\
    god this takes on it's &i was like wow & couple weeks and i &and said oh well\\
    like nasa now we got to &this is awesome i &was trying to keep & it's a waste of\\
    like oh how are we gonna &mean you can do it & my cool so i sat& money but i think\\
    do it ok and we hunker & again but i want& on the floor next&we can do it anyway\\
    down we crouch on the &you to try to get &to the toilet and &i take a deep breath\\
    ground and where's the & out of the car& waited i had my&and open the door\\
    wind coming from we're & before the guy comes& hand in the bowl &a little bit so i \\
    stopping i take out my & back you can tell&when he came to &can get my head\\
    cigarettes let's get the &him to turn off & pick it up he said&up the guy comes\\
    cigarettes ready oh my & the lights but you&you can't use it &over to me and says\\
    brand she says not & don't know how much&because i know how  &you can do it\\
    surprising and & it & it feels & i don't have any \\
    
    \hline
    we both have our & will hurt so he& so i took it&experience with this\\
    cigarettes at the ready &starts yelling at me & out of the bag&so he pulls out\\
    she strikes once nothing & i do it again and& and it started to& the knife and cuts\\
    she strikes again yes & again until my face& bubble and then popped&me up i feel the\\
    fire puff inhale mm &is bloody i end & out i grabbed my &hot blood flow into\\
    sweet kiss of that &up sitting down with & hand and held it&my mouth and then\\
    cigarette and we sit &him and just looking & out to her while&i just sit there\\
    there and we're loving &at him for the &she took the bottle & and try to suck it \\
    the nicotine and we & first time i remember& i didn't want to&out i felt so \\
    both need this right &how scared he was &take it because it &sick and i knew\\
    now i can tell the & when we were in& was so weird to&that my life was\\
    night's been tough &his room we sat & me after about minutes&about to end i\\
    immediately we start & & she&went back inside and\\
    to reminisce &  &  & started \\
    
    \hline
    about our thirty second & there for about minutes& started to cry and&talking to my friends\\
    relationship i didn't & talking about it then & i said that i wasn't&about how horrible the\\
    think that was gonna &he starts saying things &ready to see her &day was and how\\
    happen me neither oh &like you don't know & i felt so bad and&i couldn't believe it\\
    man that was close oh &who i am and what & she was like oh&but i was glad\\
    i'm so lucky i saw you &happened and then i & my god what happened &that he came over\\
    yeah then she surprises & hear this guy ask&and then she asked &i wanted to tell\\
    me by saying what was &me if i knew who & me why i was so&him about it and\\
    the fight about and i & he was i said no&upset i said i don't &he asked me what\\
    say wha what are they & i didn't and he says& know what you mean&happened he then said\\
    all about and she said & you didn't know him&but it is obvious & i don't know what\\
    i know what you mean &did you you &that you are angry &you mean what did\\
    she said was it a bad one & & &i do to him\\
    and and i said &  &  & and i replied \\
    
    \hline
    you know like medium & think i did and& about something i asked&oh it was nothing\\
    she said oh and we start & then we start talking& what happened and he &he just went on\\
    to trade stories about & about our mutual friends&said he went to &to tell me about\\
    our lives we're both from & we went to the same& visit a friend in&his childhood in the \\
    up north we're both kind & school but she lived&a town i lived in &united states he was\\
    of newish to the neighborhood &in a city about & in the south i didn't&from the east coast\\
    this is in florida we both &hours away from where & really know where it&he had lived there for\\
    went to college not great & we were she had a& was he came back &years and was an \\
    colleges but man we &great job but i & to the states to&excellent student at a \\
    graduated and i'm actually &don't see her much &attend university he said &university he also had\\
    finding myself a little & as a manager but& that he was a very &an engineering degree but\\
    jealous of her because &  she is really close& smart person but he&it didn't compare to\\
    she has this really cool & to the other people & didn't think he & mine my mom had \\
    &who work at this&would get a job as&a great job she\\
    &&an assistant to&\\
    \hline
\end{tabular}

\hspace{-3em}
\begin{tabular}{|c|c|c|c|}
    \hline
    \textbf{Ground Truth} & \textbf{S01} & \textbf{S02} & \textbf{S03} \\
    \hline
    job washing dogs she had &company i know this  &an engineering professor &worked as a secretary\\
    horses back home and &because she had to &the thing is he &for a bank for years\\
    she really loves animals & leave school because of& really doesn't want&and she has a degree\\
    and she wants to be a vet &it but she said she &to work at a university &in accounting but it's\\
    and i'm like man you're & wanted to be a teacher&so i think he can &just not possible i\\
    halfway there i'm a waitress &i got a job at & only do so much but&could work at a bank\\
    at an ice cream parlor so &a local coffee shop  &he has an engineering &but i don't want\\
    um that's not i don't know & i didn't like it& degree which i don't&to go back and\\
    where i want to be but i & but i was happy& know what that means&do that it seems\\
    know it's not that and &there and i did it &for me so i think &like you need to\\
    then it gets a little & anyways one day the& i need to start talking&\\ 
    deeper {cg} & guy comes & to people &  \\
    \hline
    and we share some other & in with a bunch of&i can trust who &talk about what happened\\
    stuff about what our lives & people who were already&don't really know &i think it is\\
    are like things that i &in the room i didn't & me but i want to&important to explain\\
    can't ever tell people at & know them well so& share what happened with&some things and maybe\\
    home this girl i can tell &i couldn't tell them &them i feel that &i can help you get\\
    her the really ugly stuff & what happened but they& if they don't like&a perspective on how\\
    and she still understands & didn't really want to&me that they can't  &you feel in your\\
    how it can still be pretty &hear the story because & be trusted to not&own way i think it\\
    she understands like how & they knew how much i& judge me for something&would be easier to\\
    nice he's gonna be when &hated the fact that& i did wrong but if &understand it myself if\\
    i get home and how & i didn't get to&i try to fix it  &i understood how i felt\\
    sweet that'll be & see him because & they will always hate & when i was younger my \\
    &he didn't care and&me because i do it again&parents had no clue\\
    &&&how much it\\
    \hline
    we are chain smoking & i couldn't stop crying& i can go to a bar&affected me so i\\
    off each other oh that's &because of it so i  &and drink some beer &decided to give them\\
    almost out come on and & left it on the counter& but they will ask&a big hug and kiss\\
    we we go through this &for him to take it &me to leave after i &their cheek then they\\
    entire pack until it's & off and he did and& finish up i walk&pulled back and started\\
    gone and then i say you & then i said i don't& outside and say i&crying i said okay\\
    know what uh this is a &want to keep it & need to get a ride&that was it i didn't\\
    little funny but you're & because i can't afford& home because i don't&have time to give\\
    gonna have to show me &to get it back & know how to drive&them any of my money\\
    the way to get home &because he didn't buy & in my state and i&i don't have a car\\
    because although i'm & me a ticket i had&didn't have my license &i can't even get a\\
    twenty three years old &no idea where to & yet so i asked him &license because of\\
    i don't have my driver's &go so i asked him &if he could pick &this so i just\\
    license yet and i just & if i could& me up to go&drove by and asked\\
    jumped out right when & & &if she could\\
    i needed to and she says & & &\\
    well why don't you come back &  &  &  \\
    
    \hline
    to my house and i'll &stay with him at & home he said sure i&drive me home she \\
    give you a ride i say & my place he said sure&  would like that so&says sure and we\\
    ok great and we start &and we walked outside & we drive out of&go and i was like\\
    walking and uh we get & together we started talking& town and down a main&ok so we get in\\
    to this um lots of uh & and then i noticed that& road where it becomes&the car and she starts\\
    lights and uh the roads & there were people & more rural the houses&driving to the address\\
    are getting wider and wider & everywhere the whole street &get smaller and less &i give her the directions\\
    and there's more cars and & was empty we looked up & populated i see a few&to her house and\\
    i see um lots of stores & and saw the sky was lit&farms and then a &then we head over there\\
    you know laundromats and &up with orange and & big town with lots&we walk through the\\
    dollar stores and emerge centers & yellow and then a car came & of stores and & neighborhood and see a \\
    &&&police car with\\
    \hline
\end{tabular}

\hspace{-3em}
\begin{tabular}{|c|c|c|c|}
    \hline
    \textbf{Ground Truth} & \textbf{S01} & \textbf{S02} & \textbf{S03} \\
    \hline
    and then we cross over & around the corner and&restaurants i drive &a red light on \\
    us one and uh she leads & we watched it until& about a mile up a &the corner in the\\
    me to some place and i & it pulled up to us&hill and i see this &street next to my\\
    think no but yes carl's & and i thought oh my&sign that reads i &house i saw this guy\\
    efficiency apartments & god it was the guy& don't like this place&and i freaked out\\
    this girl lives there and &who ran the store i & but i also think&because i thought he was\\
    it's horrible and it's & remember this i was& that the owner is a&my neighbor who had\\
    lit up so bright just to & so freaked out because& complete idiot because&a huge house that\\
    illuminate the horribleness & i knew that it was& he has a huge house&was really nice but\\
    of it it's the kind of &a bad thing because &and the whole place &was not the kind\\
    place where you drive your & of what & is built into the building & of place you would see \\
    &  &  & in a movie where everyone \\
    
    \hline
    car right up and the & happened the guy was& so you can walk in&was living together\\
    door's right there and & on the couch with a& the door without touching&but you couldn't go\\
    there's fifty million &huge amount of alcohol &the wall the windows &in without a security\\
    cigarette butts outside &and the room was full & are all covered and&camera in the front room \\
    and there's like doors one & of people who were& there are a bunch of&the tv is always on\\
    through seven and you & clearly drunk the whole&people in there that &and people are constantly\\
    know behind every single &thing was pretty weird & are probably doing nothing&talking and laughing and\\
    door there's some horrible & as the group was&but drinking or just &all sorts of weird stuff\\
    misery going on there's &drinking and smoking weed & sitting around waiting for&going on i can't\\
    someone crying or drunk & i didn't know what was&someone to call so & believe this is happening\\
    or lonely or cruel and i &going on so i went &i'm like man i feel  &i didn't know he\\
    think oh god she lives & back inside and called& bad for him i really &was gay so i went\\
    here how awful we go to & the cops i had to&do i was trying  &to my room and\\
    the door door number four &walk out & to tell him&\\
    and she very &  &  & \\
    \hline
    very quietly keys in &of there and leave the & to leave when the&was just standing there\\
    as soon as the door opens & door open because i& door opened and i saw&trying to calm myself\\
    i hear the blare of &could see a man & him walking towards me&down i could hear the \\
    television come out and & in his room in a& i could see that the&water running in the\\
    on the blue light of the & suit that was covered& light from the kitchen &bathroom and the tv\\
    television the smoke of a &in blood and he had &was off and his face &was off the only light\\
    hundred cigarettes in that &a huge knife sticking & was covered in blood&came from the screen it\\
    little crack of light and &out of his chest i & but i had never seen&was dark outside i\\
    i hear the man and he says & remember him saying oh& him before so i said&heard her say you have\\
    where were you and she says &you can go now but &hey man what are &to go now but i \\
    never mind i'm back and & i told him i don't& you doing here you need&said ok i don't\\
    he says you alright & want to he said & to go home he says i & think so and then she \\
    \hline
    and she says yeah i'm &ok but you can stay & don't know but i&was like oh okay\\
    alright and then she turns &i walked over to my &want to talk to you &you can come with\\
    to me and says you want &mom and she told me & i told him no but&me she then asked if\\
    a beer and he says who & not to tell him to& then i see him look&i was sure about my\\
    the fuck is that and she & come over he said he&at me and i think &mom i said yes and\\
    pulls me over and he sees &didn't like it i &oh i can't believe &she was like oh\\
    me and he says oh hey i'm &had a few beers and &this he must be &this is great so\\
    not a threat just then he & started to get really& crazy i was staring&i took it out of his\\
    takes a drag of his & drunk but the alcohol& at his hand and it&mouth and it was really\\
    cigarette a very hard hard &  was too much for& started moving a little&hard to swallow because\\
    drag you know the kind & my head to handle and&faster than normal it &it felt like someone\\
    that makes the end of it & & was like the weirdest&was pushing\\
    really heat up hot hot hot &  &  &  \\
    \hline
\end{tabular}

\hspace{-3em}
\begin{tabular}{|c|c|c|c|}
    \hline
    \textbf{Ground Truth} & \textbf{S01} & \textbf{S02} & \textbf{S03} \\
    \hline
    and long and it's a & i could only hold the&feeling you get when &my throat down and\\
    little scary and i follow &glass up to the light & you touch your arm  & then sucking it out \\
    the cigarette down because & to see what i had done&but then you see the  &then the water started\\
    i'm afraid of that head & wrong but it was all&blood flow down the &flowing back up into\\
    falling off and i'm & over my hand the blood& front of your body and &my body i didn't\\
    surprised when i see in &was still running down & you can feel your hand&realize what was happening\\
    the crook of his arm a &my arm the nurse saw & on the ground in a&until my hand was\\
    little boy sleeping a &  that and tried to pull&way you never imagined &touching the bottom of\\
    toddler and i think and &him away she then grabbed & possible so i went over&my leg and the other\\
    just then the girl reaches & the iv bag and began & to his car and grabbed&was resting on the\\
    underneath the bed and &to & a paper bag full&seat behind me he then\\
    takes out a carton and & & &pulled a small bottle\\
    she taps out the &  &  & out of the glove box \\
    
    \hline
    last s pack of cigarettes & fill it up with water&of food from his kitchen i &and poured some water in\\
    in there and on the way &as i ran back to the & then walked around to& it then placed it next\\
    up she kisses the little & room i looked at her and&the front door to open &to me as he sat back\\
    boy and then she kisses &said something like hey & it and he asked me what&down he said hey buddy \\
    the man and the man says & what are you doing here&i wanted and i said just &you wanna come with me\\
    again you alright and she &i'm gonna go get some & some food i went and &i replied yeah i want to\\
    says yeah i'm just gonna & food so i left the room& got a few plates and& play this game we went\\
    go out and smoke with her & and started walking down& then we started to eat&on a tour of the town\\
    and so we go outside and & the street and i thought& the meal we had already&and we were talking about\\
    sit amongst the cigarette &hey man i don't know &paid for and the guy &what kind of music the city\\
    butts and smoke and i say &you i mean i think & was like wow dude that&was i said wow this is\\
    wow that's your little boy & you're really& is awesome&amazing and\\
    and she says yeah isn't he & & &\\
    beautiful and i &  &  &  \\
    
    \hline
    say yeah he is he is & weird but you could be& i think he is really&he replied yeah it was\\
    beautiful he's my light & the best person you can&cool i mean i was and  &so much fun he was a\\
    he keeps me going she & be you don't know how& so was the other guy we&really great guy but he\\
    says we finish our &to do this but i was & went out to dinner and&didn't have any money\\
    cigarettes she finishes & told to go out and get& i didn't drink much because&he wasn't allowed to smoke\\
    her beer i don't have & drunk with friends i didn't&i was worried i would get &weed because he was too\\
    a beer cause i can't go &do it and then the guy &arrested or something like &scared of getting arrested\\
    home with beer on my &came back into my room & that the police came and&and then one night i came\\
    breath and she goes & & &home drunk and\\
    inside to get the keys & & &\\
    she takes too &  &  &  \\
    
    \hline
    long in there getting & and i started to freak& searched my house but i&it was dark i didn't\\
    the keys and i think & out thinking something& didn't hear them come&feel like talking to him\\
    something must be wrong &was wrong but then i &back they said they couldn't &so i just left he said i\\
    and she comes out and & remember him saying oh&find the keys to the &can't go with you because\\
    she says look i'm really &hey you know i don't &car and i was scared & my dad doesn't have any\\
    sorry but um like we & have any money to buy&because they didn't know &money for the bus but\\
    don't have any gas in & a new car i think&the code to get into &that i could take a cab\\
    the car it's already on &he said i can't do that & the house so i said okay&home or something so he told\\
    e and he needs to get to &so i just go with him & i can drive but i need&me to drive in the\\
    work in the morning and & and tell him to do& a map first so he shows&direction of my house which\\
    um i you know i i'm gonna &whatever it takes he then &me where it is he says &is about miles away and\\
    be walk to work as it is & says well you can try& it's about minutes away&i should head\\
    so what i did was though &to find the place where & so we walk down&\\
    here look i drew out this & & &\\
    map for you and you're & & &\\
    really you're like a mile & & &\\
    and a half from home and & & &\\
    um if you walk &  &  & \\
    
    \hline
\end{tabular}

\hspace{-3em}
\begin{tabular}{|c|c|c|c|}
    \hline
    \textbf{Ground Truth} & \textbf{S01} & \textbf{S02} & \textbf{S03} \\
    \hline
    three streets over & they are supposed to& the road to where he&down the road a little\\
    you'll be back on that & meet and then i go in& lives and i take out a &ways until i see the\\
    pretty street and you just & there and he asks me& small piece of paper and&sign and pull over and\\
    take that and you'll be & to bring a small bag& write this on the inside&park the car the kid\\
    fine and she also has &with a few bottles of & of the windshield with the&comes out with a bag of\\
    wrapped up in toilet & wine in it i said ok& red pen that she gave&chips and a small box\\
    paper seven cigarettes & and took them inside and& to me the whole thing&of candy i take them from\\
    for me a third of her pack & we started talking and&was really cool and she &him and he says hey man\\
    i note and a new pack of &he asked me about my & asked me what i wanted&i'm so glad i got\\
    matches and she tells me & life and how i felt& for Christmas she told&to meet you i like you\\
    good bye and that was & about him and& &too and\\
    great to meet you and & me to& &\\
    how lucky and that & & &\\
    was fun and you know &  &  & \\
    
    \hline
    let's be friends and i & i said yes and then& call her anytime and&then he said ok well\\
    say yeah ok and i walk & i left because i wasn't&i said sure and hung &that was weird i guess i \\
    away but i kind of know & sure what was happening &up i didn't really know &just didn't want to go\\
    we're not gonna be friends & but i wanted to see her&what to do but i thought &in there and he just\\
    i might not ever see her & again it was really sad& she would be okay i& kind of left after that \\
    again and i kind of know & when i realized she& don't think she ever talked&but i think it was really\\
    i don't think she's ever & didn't even know i was& about it i didn't really&interesting for me because i\\
    going to be a vet and i & there until after i& know what was happening&had always wanted to be a\\
    cross and i walk away and &left i went home and & until i went to see her&doctor but it seemed like\\
    maybe this would've seemed &sat down to think i & after that and it was&they had a different \\
    like a visit from my & felt a & hard to process & approach than what \\
    
    \hline
    possible future and scary & wave of panic that i& because i was so confused&i was used to in high \\
    but it kind of does the & had never experienced&  and angry but then i&it was really hard \\
    opposite on the walk home &  before but i didn't really& started talking to my&to understand what was \\
    i'm like man that was & feel like it was happening& parents about what they&happening but it seemed\\
    really grim over there & so i said hey man& had done and about how&like there were people\\
    and i'm going home now & you know this guy that& my sister was a horrible&all around us i noticed\\
    to my nice boyfriend and & came back to me after& person and how i didn't&one guy who was very\\
    he is gonna be so extra & i went to bed and was& know it then the worst &big and scary looking but\\
    happy to see me and we & really depressed he had& thing ever happened she&he was also a good friend\\
    have a one bedroom & been drinking for months& was living in a tiny&and he had a small\\
    apartment and we have & he was living in an& apartment with a lot of&apartment which he shared\\
    two trees and there's & apartment with two other & stuff in & with two other guys it was a \\
     & guys and it &  &  \\
    \hline
     a yard and we have & was a huge house with a big & her bedroom and a kitchen & huge room with a tv\\
     this jar in the kitchen & screen tv and a couch so & that had no electricity& in the corner and\\
     where there's like & he could play video games & i had to pay for everything& a kitchen that was\\
     loose money that we & i didn't have any money for& myself i didn't really& filled with food that\\
     can use for anything & anything i had no friends & have any money saved up&he kept on hand i\\
     like we would never & so i just stayed there alone & but my mom would have&didn't have any money\\
     ever run out of gas & until he left i started drinking & helped me out but i & because i wasn't paying\\
     and um i don't have a & a lot and i didn't really & wasn't going to do&for anything and my\\
     baby you know so i can & know what was happening & anything i just stayed&dad didn't want me to\\
     leave whenever i want & until my friends found & with him for about two&go out drinking so i\\
     i smoked all seven & out and they didn't & days until he stopped&just walked home that\\
     cigarettes on the way & believe me & drinking i didn't know& night after work he said\\
     home and people who have & & how much he had had& that i needed to stop\\
     never smoked cigarettes & & but he said&smoking weed because \\
     just think ick &  &  &  i didn't\\    

    \hline
\end{tabular}

\hspace{-3em}
\begin{tabular}{|c|c|c|c|}
    \hline
    \textbf{Ground Truth} & \textbf{S01} & \textbf{S02} & \textbf{S03} \\
    \hline
    disgusting and poison & when i said it i couldn't & that it was good to& like the taste it was\\
    but unless you've had & even explain it but i & feel like you are& disgusting and it made\\
    them and held them dear & think it's important& not alone i think it&me feel sick but it\\
    you don't know how great & that you know why it& is important to look& also helped me to be\\
    they can be and what & happened you can do it& back on this moment& more comfortable around\\
    friends and comfort and & because you are so powerful& with some sadness but& my friends who i was\\
    kinship they can bring & you have an ability to help&  it will come to a & close to and i found\\
    it took me a long time & other people but you must not&  head soon i feel that& a way to connect with \\
    to quit that boyfriend & give up you need to learn& we need to change the& 
 them after the breakup i \\
    and then to quit & to accept what happened& past so we can live& think they got over it\\
      & and move on i had a & it again we do it &  \\
    
    \hline
    smoking and uh & girlfriend when i was& now it was a mistake & but i was still depressed\\
    sometimes i still & and a year later she& but the guy is a& for months after she died\\
    miss the smoking & divorced my
    &  & \\
    \hline
\end{tabular}


\end{document}